%% file: main.tex
\pdfoutput=1
\documentclass[letter]{article}

    \usepackage[nonatbib, preprint]{neurips_2023}

\usepackage[utf8]{inputenc} 
\usepackage[T1]{fontenc}    
\usepackage{hyperref}       
\usepackage{url}            
\usepackage{booktabs}       
\usepackage{amsfonts}       
\usepackage{nicefrac}       
\usepackage{microtype}      
\usepackage{xcolor}         
\usepackage{amsmath}
\usepackage{caption}
\usepackage{subcaption}
\usepackage[pdftex]{graphicx}
\usepackage{enumitem}
\usepackage[sort&compress]{natbib}

\DeclareMathOperator*{\argmin}{argmin}

\title{Neural Network Reduction with Guided Regularizers}

\author{%
  Ali Haisam Muhammad Rafid \\
  Department of Computer Science\\
  Virginia Tech\\
  Blacksburg, VA 24061 \\
  \texttt{haisamrafid@vt.edu} \\
   \And
  Adrian Sandu \\
  Department of Computer Science\\
  Virginia Tech\\
  Blacksburg, VA 24061 \\
  \texttt{sandu@cs.vt.edu} \\
}

\begin{document}

\input{front}

\maketitle

\begin{abstract}
\input{abstract}
\end{abstract}

\section{Introduction}
\input{introduction}

\section{Background}
\input{background}

\section{Methodology}
\input{methodology}

\section{Empirical Evaluation of Guided Regularizers}
\label{sec:empirical}
\input{empirical}

\section{Conclusions}
\input{conclusion}




\input{main.bbl}

\end{document}

%% file: front.tex
\thispagestyle{empty}
\setcounter{page}{0}

\makeatletter
\def\Year#1{%
  \def\yy@##1##2##3##4;{##3##4}%
  \expandafter\yy@#1;
}
\makeatother

\begin{Huge}
\begin{center}
Computational Science Laboratory Technical Report CSL-TR-\Year{\the\year}-{2} \\
\today
\end{center}
\end{Huge}
\vfil
\begin{huge}
\begin{center}
Ali Haisam Muhammad Rafid\\
Adrian Sandu
\end{center}
\end{huge}

\vfil
\begin{huge}
\begin{it}
\begin{center}
``Neural Network Reduction with Guided Regularizers''
\end{center}
\end{it}
\end{huge}
\vfil

\begin{large}
\begin{center}
Computational Science Laboratory \\
Computer Science Department \\
Virginia Polytechnic Institute and State University \\
Blacksburg, VA 24060 \\
Phone: (540)-231-2193 \\
Fax: (540)-231-6075 \\ 
Email: \url{sandu@cs.vt.edu} \\
Web: \url{http://csl.cs.vt.edu}
\end{center}
\end{large}

\vspace*{1cm}

\begin{tabular}{ccc}
\includegraphics[width=2.5in]{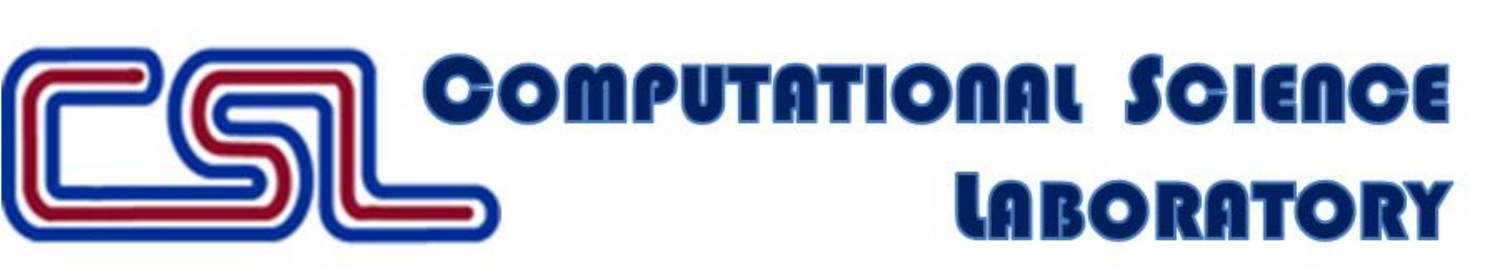}
&\hspace{2.5in}&
\includegraphics[width=2.5in]{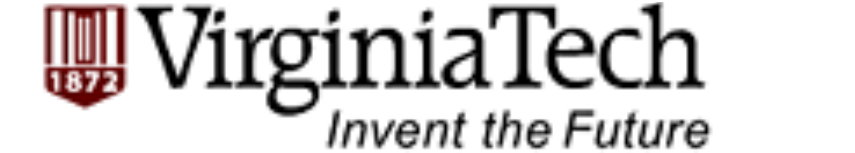} \\
{\bf\em\large Compute the Future} &&\\
\end{tabular}
\newpage

%% file: abstract.tex
Regularization techniques such as $\mathcal{L}_1$ and $\mathcal{L}_2$ regularizers are effective in sparsifying neural networks (NNs). However, to remove a certain neuron or channel in NNs, all weight elements related to that neuron or channel need to be prunable, which is not guaranteed by traditional regularization. This paper proposes a simple new approach named ``Guided Regularization'' that prioritizes the weights of certain NN units more than others during training, which renders some of the units less important and thus, prunable. This is different from the scattered sparsification of $\mathcal{L}_1$ and $\mathcal{L}_2$ regularizers where the the components of a weight matrix that are zeroed out can be located anywhere. The proposed approach offers a natural reduction of NN in the sense that a model is being trained while also neutralizing unnecessary units. We empirically demonstrate that our proposed method is effective in pruning NNs while maintaining performance.

%% file: introduction.tex

Along with the growing popularity of neural networks (NN) in machine learning, and their application to more and more complex problems in many fields, the size of NNs has been constantly increasing. Today, some  state-of-the-art models have billions of parameters to train. While the availability of powerful GPUs makes the training of such models possible, the large number of parameters also requires a large amount of memory, large power to run, and long inference times. It is important to reduce the size of NNs that can accomplish a given task in order to lower power usage and reduce inference time. For example, applications that require quick decision-making, such as collision detection, need small networks for quick inference.

Several techniques have been proposed for neural network size reduction. One of the earliest is based on approximating model parameters using quantization and hashing \cite{courbariaux2015binaryconnect,gong2014compressing}. Parameter approximation is performed using low-rank tensor factorization in \cite{denil2013predicting, lebedev2014speeding, su2018tensorial}. Knowledge distillation, discussed in  \cite{buciluǎ2006model, hinton2015distilling}, trains a small NN to copy the behavior of a large NN \cite{buciluǎ2006model, hinton2015distilling}. 

Weight pruning methods can be categorized as i) unstructured or ii) structured. Unstructured pruning methods focus on removing individual weight components instead of removing units such as neurons or channels. On the other hand, structured pruning methods remove units such as neurons, channels or attention heads. In this work, we propose a structured pruning method inspired by the unstructured pruning approach. We apply the idea of regularization, which is an unstructured approach, for removing individual weights and apply it to remove regions of weights from NNs. The approach is inspired by two classes of methods: i) sparsification of weights using regularizers during training \cite{collins2014memory, voita2019analyzing} and ii) train/prune/fine-tune framework where pre-trained neural networks are pruned by removing unnecessary units \cite{kuzmin2019taxonomy, blalock2020state, hoefler2021sparsity}. 

{\it Our Contributions.}
The popular lasso \cite{tibshirani1996regression} and ridge \cite{hoerl1970ridge} regularizers lead to irregular sparsity patterns. We propose a simple modification where the sparsification is guided such as to favor regular sparsity patterns, which helps in removing entire sets of neurons or channels, thus obtaining truly smaller NNs while preserving performance. The proposed regularizers penalize the weights of some specific units more than others, and nudge the corresponding weights to smaller values. After training using the regularized loss, we apply a selection process of neurons or channels based on how small the weight components of the weight units are. Our method is simpler and more efficient than the previous pruning methods. The process involves a simple modification to the loss function and can be immediately used within the available NN implementations. Experiments demonstrate the effectiveness of our proposed methodology.

{\it Related Work on NN Pruning.}
Most of the pruning methods proposed in the literature compute some importance score and run a selection algorithm based on that importance score. The method proposed in \cite{he2014reshaping} computes two scores for a node based on its outgoing and incoming links and selects nodes with low importance values for removal. Authors of \cite{srinivas2015data} remove a neuron based on its similarity with other neurons. They reason that if two neurons have two equal weight sets, one of them is redundant. A two-step algorithm was proposed in \cite{he2017channel} where the first step involved determining representative channels of convolutional layers and removing redundant channels, and the second step involved reconstructing remaining channel outputs with linear least squares. \cite{luo2017thinet} formulates the channel selection problem as an optimization problem and uses a greedy approach to solve the problem. \cite{li2016pruning} proposes computing the sum of absolute weight components for each filter in convolutional layers and pruning a certain number of filters with the smallest sum value.  The algorithm in \cite{molchanov2016pruning} is designed to select a subset of parameters such that the difference between the cost function before and after pruning is minimized. A neural network pruning algorithm based on the idea of the data compression approach known as coresets was proposed in \cite{mussay2019data, mussay2021data}. Authors of \cite{liebenwein2019provable} computed filter sensitivities and applied a sampling based pruning algorithm. The most recent pruning algorithm was proposed in \cite{el2022data} where the authors pose the pruning problem as a weak submodular optimization problem and apply greedy selection to solve the problem.

%% file: background.tex

This section reviews relevant background on neural networks that will be helpful for understanding our proposed methodology. The notation defined here will be used throughout the paper.

\subsection{Feed-forward Neural Network}
A simple feed-forward neural network consists of an input layer, an output layer, and several hidden layers between the input and the output. All the layer consists of units called neurons. All the neurons are interconnected between the successive layers. Given $N$ number of inputs $\mathbf{x}_i$ and outputs $\mathbf{y}_i$ where $i=1, 2, \ldots, N$; a trained neural network represents a function $f$ that maps from inputs to outputs:
\begin{align}
\label{eq:nn-function}
    \mathbf{y} = f(\mathbf{x}, \boldsymbol{\theta}).
\end{align}

The neural network learns the function $f$ by tuning the parameters $\boldsymbol{\theta}$ to get the best possible approximation of the input to output mapping based on some objective or loss function $\mathcal{L}(f(\mathbf{x}, \boldsymbol{\theta}), \mathbf{y})$:
\begin{align}
\label{eq:nn-opt}
    \boldsymbol{\theta}^* = \argmin_{\boldsymbol{\theta}} \mathcal{L}(f(\mathbf{x}, \boldsymbol{\theta}), \mathbf{y}).
\end{align}
The relation between two consecutive layers in a feed-forward neural network with $L$ layers can be written as:
\begin{align}
    \label{eq:nn-layer}
    h^\ell = g\big(W^{\ell}\, h^{\ell-1} + b^\ell\big),\quad W^{\ell} \in \mathbb{R}^{m_\ell \times m_{\ell-1}}, \quad b^{\ell} \in \mathbb{R}^{m_\ell}, \quad \ell = 1,\dots,L,
\end{align}
where $h_{\ell}$ represents the output of the $\ell$-th layer. Weight matrices $W^{\ell} \in \boldsymbol{\theta}$ and bias vectors $b^{\ell} \in \boldsymbol{\theta}$ are learnable parameters that are used to project the representation of the $(\ell-1)$-th layer to the dimension of the $\ell$-th layer. If $(\ell-1)$-th layer has $m_{\ell-1}$ neurons and $\ell$-th layer has $m_\ell$ neurons, then the weight matrix $W^\ell$ will be of dimension $m_\ell \times m_{\ell-1}$ and the bias vector will be of length $m_\ell$. Each element $W^\ell_{ij}$ for $i=1,2,\ldots,m_\ell; j=1,2,\ldots,m_{\ell-1}$ of the weight matrix $W^\ell$ represents the weight element relevant to the connection from the $j$-th neuron of the $(\ell-1)$-th layer to the $i$-th neuron of the $\ell$-th layer. The function $g(\cdot)$ is known as the activation function that is used to introduce non-linearity in the neural network.

\subsection{Convolutional Neural Network}
Convolutional neural networks (CNN),  first proposed in \cite{lecun1998gradient}, consist of layers that apply convolution operations on input images using a certain number of filters known as kernels. Application of these 2D kernels can detect patterns in images and produce feature maps. For a certain convolutional layer in the $\ell$-th layer, if the input image has $k_{\ell-1}$ channels, the number of kernels is $k_\ell$ and all the kernels are of dimension $d_1\times d_2$, then the associated weight tensor $W^\ell$ is of dimension $k_\ell\times k_{\ell-1}\times d_1\times d_2$. This weight tensor produces a feature map with $k_\ell$ channels from an input with $k_{\ell-1}$ channels. A kernel $W^\ell_{ij} \in \mathbb{R}^{d_1\times d_2}$ is associated with the $j$-th channel of the input image and the $i$-th channel of the output feature map.

%% file: methodology.tex

\subsection{Intuition and Formulation}

\begin{figure}[t]
    \centering
    \begin{subfigure}{.5\textwidth}
    \centering
    \includegraphics[scale=0.3,clip]{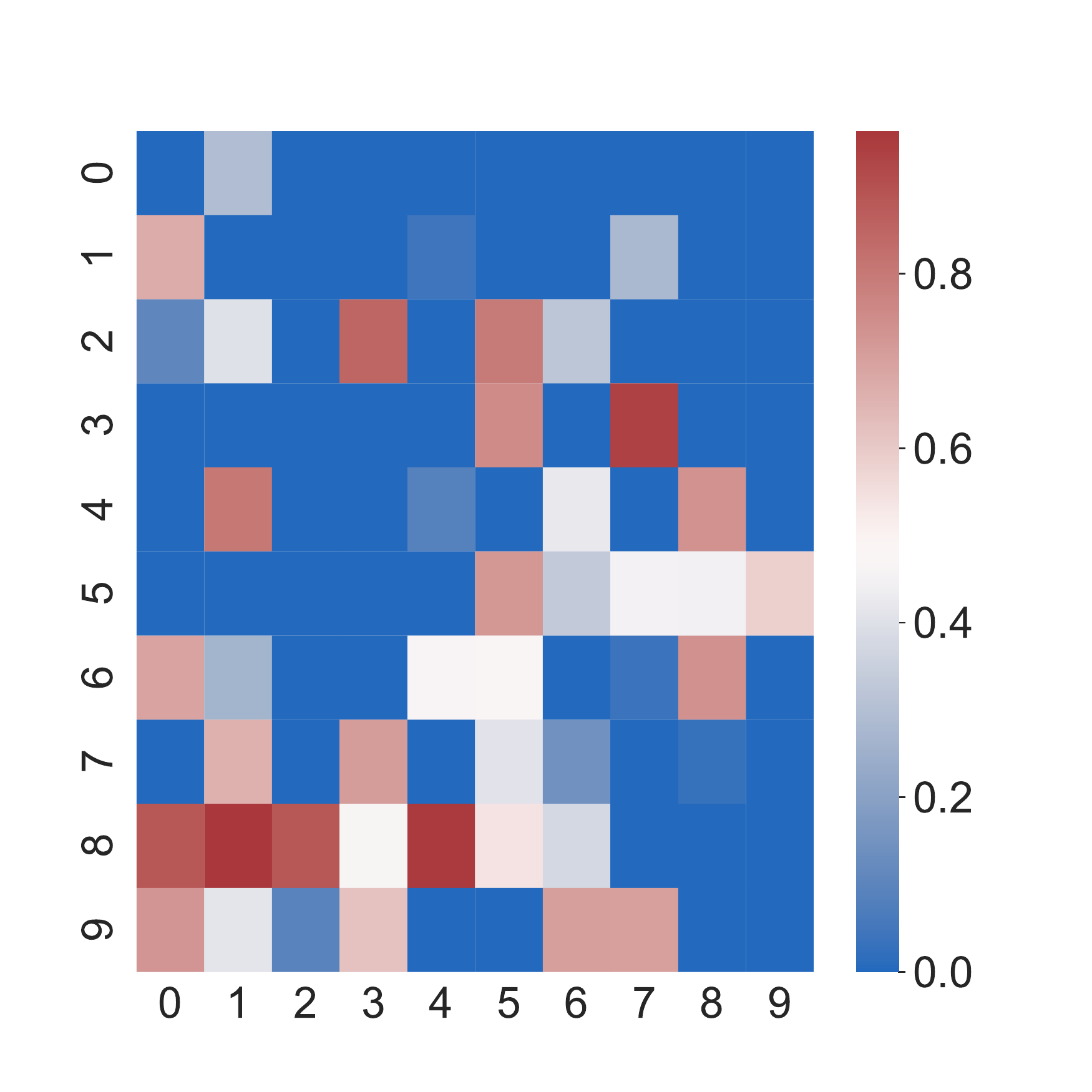}
    \caption{Standard regularization}
    \end{subfigure}%
    \begin{subfigure}{.5\textwidth}
    \centering
    \includegraphics[scale=0.3]{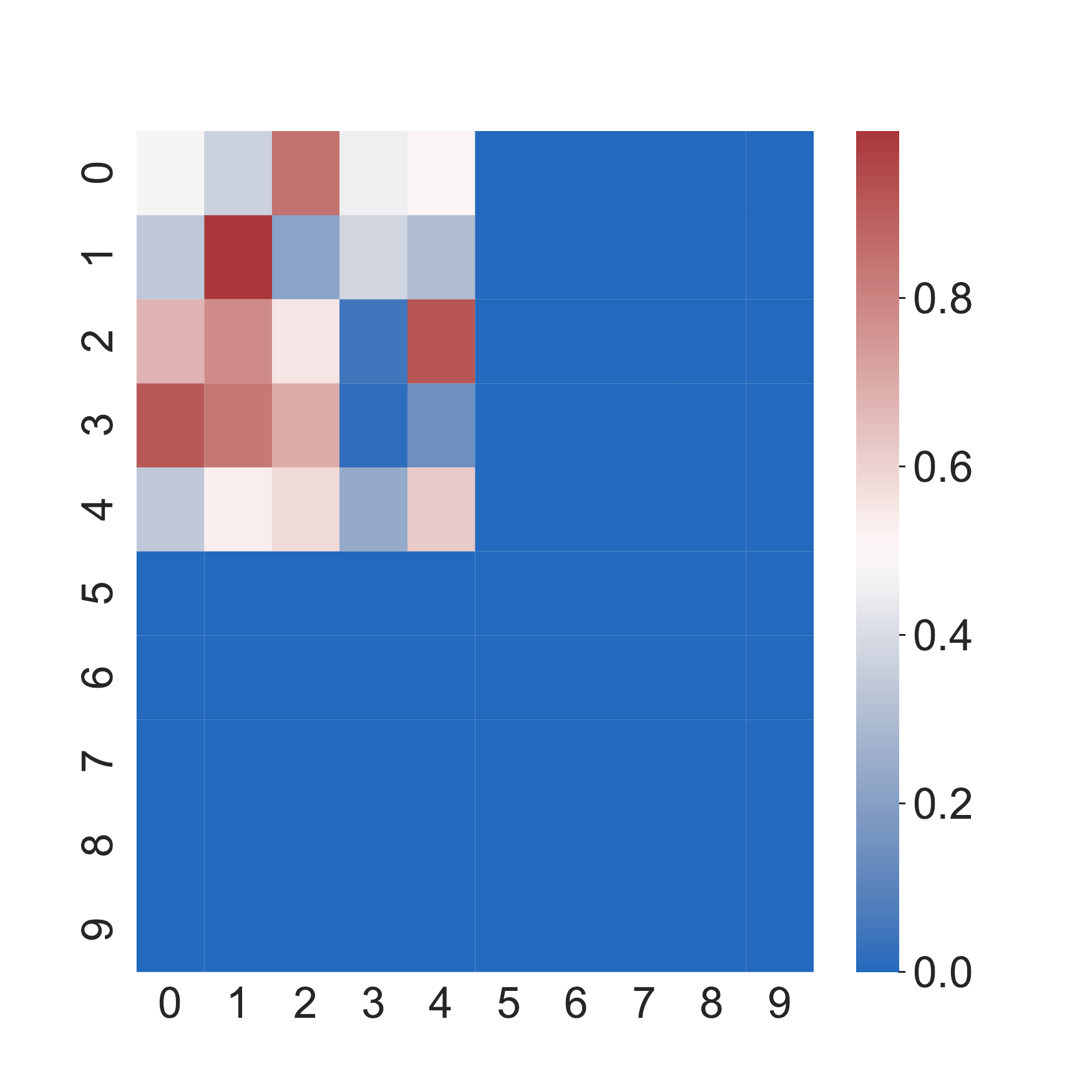}
    \caption{Guided regularization}
    \end{subfigure}
    \caption{Cartoon of sparsity patterns resulting from different regularizations for a $10 \times 10$ weight matrix; colors represent weight values. The weight matrix on the left has a scattered sparsity pattern as typically results from $\mathcal{L}_1$ or $\mathcal{L}_2$ regularizers. Although many of the weight values are zero (or are small and can be pruned to zero), one cannot eliminate individual neurons. The weight matrix on the right has a structured sparsity structure that allows to completely remove the last 5 neurons from the current layer as well as the last 5 neurons from the previous layer.}
    \label{fig:weight_example}
\end{figure}

Our methodology is based on the ideas of $\mathcal{L}_1$ regularizers \cite{tibshirani1996regression} and $\mathcal{L}_2$ regularizers \cite{hoerl1970ridge}. With regularization, the neural network training optimizes a modified loss function $\mathcal{L'}$ consisting of the traditional data loss component $\mathcal{L}$ plus a term that penalizes the magnitude of the weights ($W^{\ell} \in \mathbb{R}^{m_\ell \times m_{\ell-1}}$):
\begin{subequations}
\label{eq:regularizers}
\begin{align}
\label{eq:l1reqularizer}
    \text{Standard $\mathcal{L}_1$ regularizer:}\quad \mathcal{L'} &= \mathcal{L} + \lambda\, \sum_{i=1}^{m_\ell}\sum_{j=1}^{m_{\ell-1}} |W^{\ell}_{ij}|\\
    \label{eq:l2reqularizer}
     \text{Standard $\mathcal{L}_2$ regularizer:}\quad \mathcal{L'} &= \mathcal{L} + \lambda\, \sum_{i=1}^{m_\ell}\sum_{j=1}^{m_{\ell-1}} {W^{\ell}_{ij}}^2.
\end{align}
\end{subequations}
The regularizer term related to the weight component of each layer is added over all layers. The strength of the regularization term is tuned by the penalty factor $\lambda$. Regularizers force the neural network to push the weight values toward zero. However, this sparsity is scattered in most cases. As all the weight elements in the regularization term are given the same priority $\lambda$, the weight values that are pushed towards zero can appear anywhere in the weight matrix. But, as a row of a weight matrix contains all the weight elements for a neuron connected to all the neurons of the previous layer, being able to push all the values of a row toward zero means we can eliminate the neuron associated with the row. In the same way, being able to push all the values of a column toward zero means we can eliminate the neuron from the previous layer associated with that column. This idea is illustrated in Figure \ref{fig:weight_example}. The resulting sparsity structures from real experiments are discussed in Section \ref{sec:empirical}.

To guide the NN training toward a solution where the rows or columns of the weight matrices are nudged toward zero, we propose the following regularizers where each weight is penalized differently, depending on its location in the weight matrix:
\begin{subequations}
\label{eq:mod-reqularizer}
\begin{align}
\label{eq:mod-l1reqularizer}
    \text{Guided }\mathcal{L}_1\text{ regularizer:}\quad \mathcal{L'} &= \mathcal{L} + \lambda \sum_{i=1}^{m_\ell}\sum_{j=1}^{m_{\ell-1}} \frac{i + j}{m_\ell + m_{\ell-1}}|W^{\ell}_{ij}|\\
    \label{eq:mod-l2reqularizer}
    \text{Guided }\mathcal{L}_2\text{ regularizer:}\quad \mathcal{L'} &= \mathcal{L} + \lambda \sum_{i=1}^{m_\ell}\sum_{j=1}^{m_{\ell-1}} \frac{i + j}{m_\ell + m_{\ell-1}} {W^{\ell}_{ij}}^2
\end{align}
\end{subequations}
As the value of row index $i$ or column index $j$ increases in the formulation \eqref{eq:mod-reqularizer}, the weight components $W^{\ell}_{ij}$ are penalized more. Weight components from rows or columns with a higher index value are penalized more than those from rows or columns with a lower index; the weights in the bottom right corner of the weight matrix are penalized the heaviest, and those in the upper left corner the lightest. This formulation can of course be changed to penalize the rows or the columns with the lower index values more than the higher index values. In this paper, however, we only work with the discussed formulation.

In order to favor the pruning of neurons from only one layer, one could penalize the weight components based only on the row indices ($i/m_\ell$) or on the column indices ($j/m_{\ell-1}$) instead of using both indices. However, instead of the full network, if we just apply the regularizer to a weight matrix between two layers, this allows us to eliminate neurons from both layers. This helps with efficiently pruning the full network (see Sec. \ref{subsec:prune-multiple}).

For a convolutional layer in the $\ell$-th layer, we can think of a weight tensor $W^\ell \in \mathbb{R}^{k_\ell\times k_{\ell - 1}\times d_1\times d_2}$ as weight matrices of dimension $k_\ell\times k_{\ell - 1}$ with all the kernels of dimension $d_1\times d_2$ as elements of the weight matrix. Following the analogy of feed-forward networks, the elements of a row can be considered to be all the kernels associated with a specific channel in the output feature map and the elements of a column can be considered to be all the kernels associated with a channel in the input image. Based on these considerations, we propose the following guided regularizer formulations for convolutional layers:
\begin{subequations}
\label{eq:mod-reqularizer-conv}
\begin{align}
    \label{eq:mod-l1reqularizer-conv}
    \text{Guided }\mathcal{L}_1\text{ Regularizer: } \mathcal{L'} &= \mathcal{L} + \lambda \sum_{i=1}^{k_\ell}\sum_{j=1}^{k_{\ell - 1}} \frac{i + j}{k_\ell + k_{\ell - 1}}\|W^{\ell}_{ij}\|_1\\
    \label{eq:mod-l2reqularizer-conv}
    \text{Guided }\mathcal{L}_2\text{ Regularizer: } \mathcal{L'} &= \mathcal{L} + \lambda \sum_{i=1}^{k_\ell}\sum_{j=1}^{k_{\ell - 1}} \frac{i + j}{k_\ell + k_{\ell - 1}} \|W^{\ell}_{ij}\|_2^2
\end{align}
\end{subequations}
In this case, we substitute the weight elements with norms of kernels. Channels with higher index values for both input and output maps are penalized more. Being able to push the norms of all the kernels associated with a channel will enable us to prune entire channels along with the associated kernels. This idea is similar to the one discussed in \cite{li2016pruning}, where the authors pre-trained the model and then computed the norms of the kernels to make decisions based on the kernel norms. We are incorporating this into the loss function. The authors of \cite{li2016pruning} only removed kernels with low norms. We want to remove a channel completely.

\subsection{Pruning Condition}

To decide which neurons/channels to prune, we calculate a threshold value ($\tau$) based on the values of weight components/kernel norms in the weight matrices/tensors. For some weight matrix/tensor $W^\ell$, we compute the maximum absolute row sum $\eta_{\text{max}}$, then calculate $\tau^\ell$ for layer-$\ell$ as follows:
\begin{subequations}
    \begin{align}
    \label{eq:eta-max1}
    &\text{For weight matrix: } \eta_{\text{max}} = \max \{ \sum_{j=1}^{m_{\ell-1}} W^\ell_{ij}: i = 1, \ldots,m_\ell\} \\
    \label{eq:eta-max2}
    &\text{For weight tensor: } \eta_{\text{max}} = \max \{ \sum_{j=1}^{k_{\ell-1}} \|W^{\ell}_{ij}\|_1: i = 1, \ldots,k_\ell\} \\
    \label{eq:threshold}
    &\text{Threshold: } \tau^\ell = \alpha\cdot\eta_{\text{max}}
\end{align}
\end{subequations}
We remove all the rows from a weight matrix/tensor that fails the condition in formulation \eqref{eq:thresh-cond}.
\begin{subequations}
\label{eq:thresh-cond}
\begin{align}
    &\text{For weight matrix: }  \tau^\ell > \sum_{j=1}^{m_{\ell-1}} W^\ell_{ij}; i = 1, \ldots,m_\ell\\
    &\text{For weight tensor: }  \tau^\ell > \sum_{j=1}^{k_{\ell-1}} \|W^{\ell}_{ij}\|_1; i = 1, \ldots,k_\ell
\end{align}    
\end{subequations}
Here, $\alpha \in [0, 1]$ is a hyperparameter that controls the value of $\tau^\ell$ and so controls the number of neurons/channels we remove from a layer. When $\alpha=0$, we perform no pruning.

\subsection{Pruning Multiple Layers}\label{subsec:prune-multiple}
To prune multiple layers, we simply consider the weights from multiple layers in the modified loss function. This affects the weight units from the targeted layers while training the network. We prune the layers sequentially from the input layer to the output layer. We do not prune any neurons or channels from the input layer and the output layer. For some weight $W^\ell$ of layer-$\ell$, we remove rows from the weight based on the threshold condition $\tau^\ell$ \eqref{eq:thresh-cond}. While pruning $W^\ell$, we also have to eliminate columns associated with the neurons/channels we pruned in layer-$(\ell-1)$. To do this, we remove rows from the weight $W^{\ell-1}$ of layer-$(\ell-1)$ and delete corresponding column indices of $W^\ell$. For associated bias parameters of any layer, we only remove bias elements associated with deleted neurons/channels. The neuron removal leads to a reduced NN, which can be constructed as a standalone network, and inherits the remaining weight components from the pruned original NN.

In \cite{mariet2015diversity}, the authors suggested a reweighting procedure after pruning a layer. In reweighting, new weights for the layer-$(\ell + 1)$ are calculated to account for the change in layer-$\ell$ after pruning. However, we make no such adjustment to the weights after pruning. If we apply the regularizer to layer-$\ell$, certain neurons/channels of layer-$\ell$ become prunable while also training the weights of layer-$(\ell + 1)$ accordingly. That is why, we think reweighting is not required for our method.

\subsection{Method Implementation}

Our pruning method follows a sequence of steps in the train, prune, and fine-tune framework \cite{han2015learning} that can be easily implemented in modern deep learning libraries, as follows:
\begin{enumerate}[topsep=0pt,itemsep=0ex,partopsep=0ex,parsep=1ex] 
    \item Construct the guided regularizer and add it to the loss function;
    \item Train a large neural network model for some epochs for pre-training;
    \item Based on the computed threshold values for each weight matrix/tensor, reduce weight units;
    \item Define a new reduced model based on the reduced weight matrices/tensors;
    \item Load the reduced weights to the new model;
    \item Fine-tune the model by training it more for a small number of epochs.
\end{enumerate}

\subsection{Pruning Cost}
During training with regularization, computation of the regularizer term for the weight unit of layer-$\ell$ takes $\mathcal{O}(m_\ell\cdot m_{\ell-1})$ operations, where $m_\ell$ is the number of output neurons/channels and $m_{\ell-1}$ is the number of input neurons/channels. The regularizer terms can be computed in parallel for all the layers. 

During the pruning step, computation of the threshold $\tau^\ell$ \eqref{eq:threshold} for layer-$\ell$, specifically $\eta_{\text{max}}$ (\ref{eq:eta-max1}, \ref{eq:eta-max2}) takes $\mathcal{O}(m_\ell\cdot m_{\ell-1})$ operations. Pruning of a weight unit $W^\ell$ for layer-$\ell$ takes $\mathcal{O}(m_\ell + m_{\ell-1})$ operations, $m_\ell$ iterations are required for pruning rows of a weight unit and $m_{\ell-1}$ iterations are required for pruning columns. If the pruning is done for $L$ layers, the overall time complexity for pruning a neural network becomes $\mathcal{O}((m_\ell + m_{\ell-1})\cdot L)$, considering operations for the largest weight unit as the upper bound.

%% file: empirical.tex

In this section, we demonstrate the effectiveness of our proposed pruning method in several experiments. For all the experiments, we report the results after pruning the pre-trained model in a single step, followed by fine-tuning the reduced network.

\subsection{Metrics}
For each experiment, we report how the classification accuracies of the reduced NN models change while increasing the compression ratios. Compression ratio is the ratio between the original NN size and the size of the reduced NN after pruning. The $\alpha$ values \eqref{eq:threshold} are chosen from the set $\big\{0, 10^{-6}, 10^{-5}, 10^{-4}, 10^{-3}, 10^{-2}, 0.05, 0.10, 0.15, \ldots, 1.00\big\}$. For each $\alpha$ value, we prune the original network and if the desired compression ratio is achieved, we stop and report the performance. All the results are reported for the test dataset of the datasets used. 

\subsection{Experiment 1: Simple Experiment for Weight Matrix Visualization}
The first experiment illustrates how the guided regularizer affects the weight matrices in a very simple feed-forward neural network. 

\subsubsection{Experimental Setup}
We train a simple feed-forward NN with two hidden layers, consisting of 200 and 200 neurons respectively, on the MNIST dataset \cite{lecun1998gradient}. 
This model is trained for 50 epochs with a batch size of 256 and a learning rate of $10^{-3}$. It is optimized using Stochastic Gradient Descent (SGD) with a Nesterov momentum value of 0.9 during training \cite{robbins1951stochastic, nesterov1983method}. After pruning, it is fine-tuned for 5 epochs while using the Adam optimizer ($\beta_1 = 0.9$ and $\beta_2 = 0.99$) \cite{kingma2014adam}. We apply our method for $\lambda = 10^{-2}$ (the regularizer penalty) during training. For fine-tuning, we used the normal loss function. 

\subsubsection{Results and Discussion}
\begin{figure}[htb!]
    \centering
    \begin{subfigure}{.33\textwidth}
    \centering
        \includegraphics[scale=0.28]{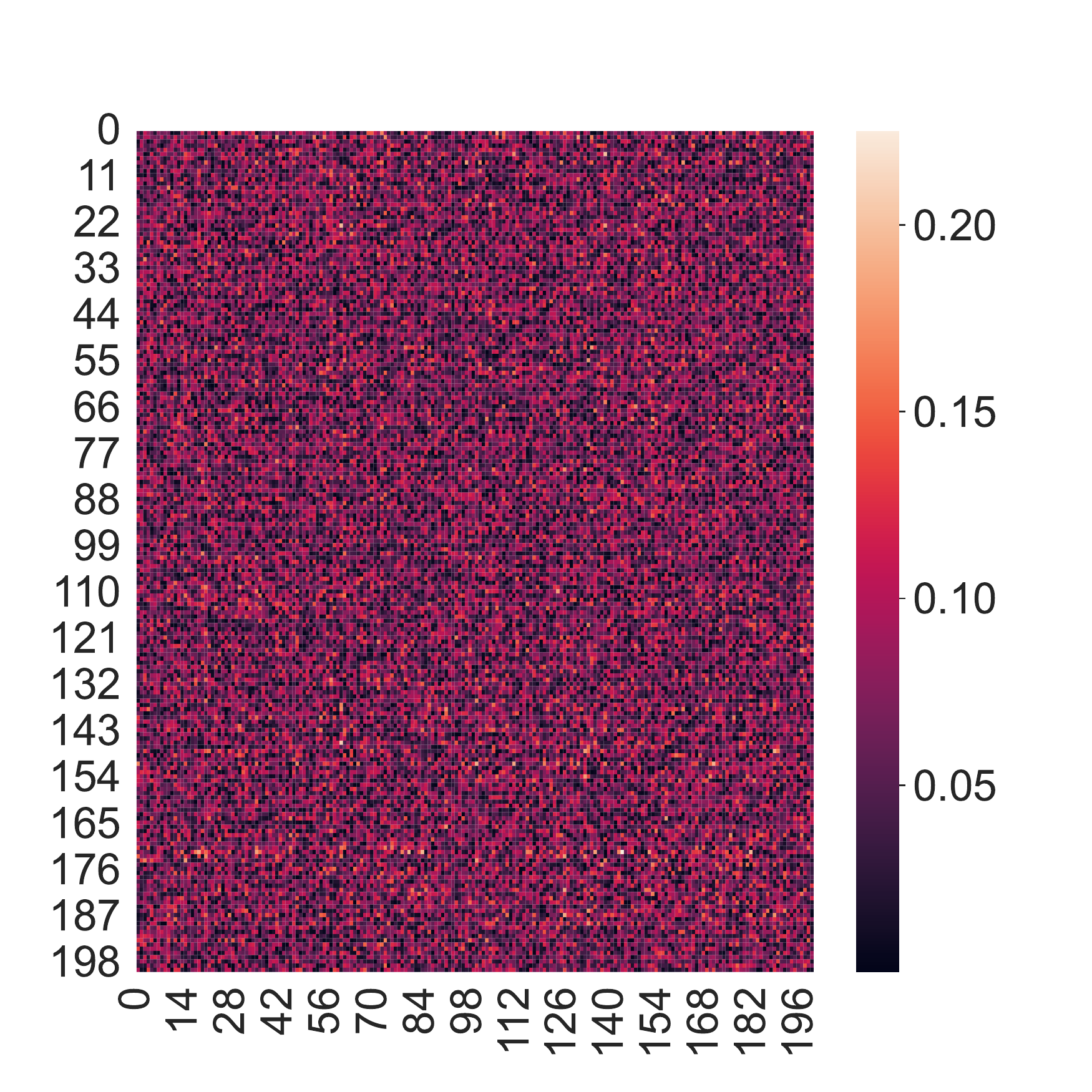}
        \caption{No regularizer}
    \label{fig:exp1-weights-fnn-none}
    \end{subfigure}%
    \begin{subfigure}{.33\textwidth}
    \centering
        \includegraphics[scale=0.28]{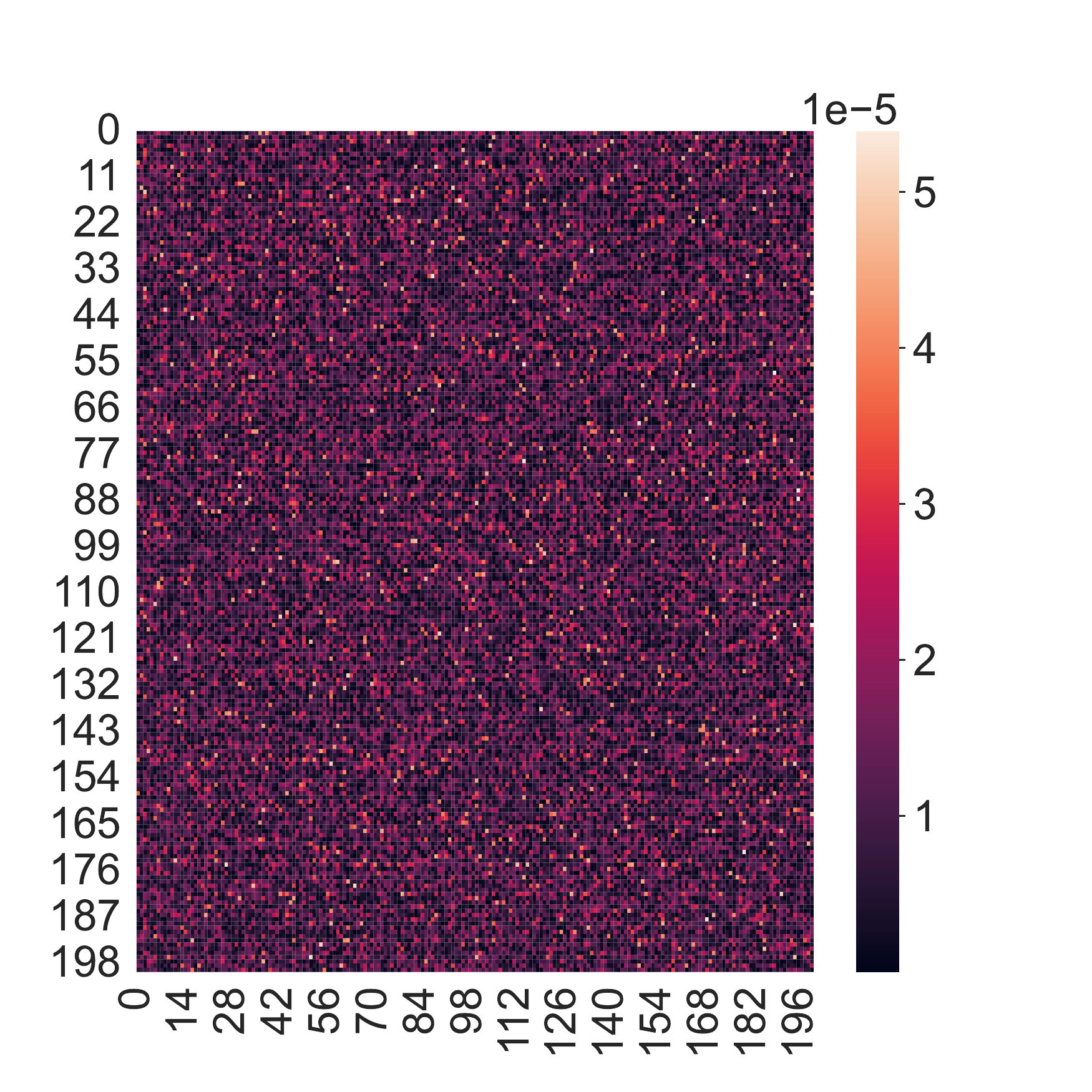}
        \caption{Standard $\mathcal{L}_1$ regularizer}
        \label{fig:exp1-weights-fnn-l1}
    \end{subfigure}%
    \begin{subfigure}{.33\textwidth}
    \centering
        \includegraphics[scale=0.28]{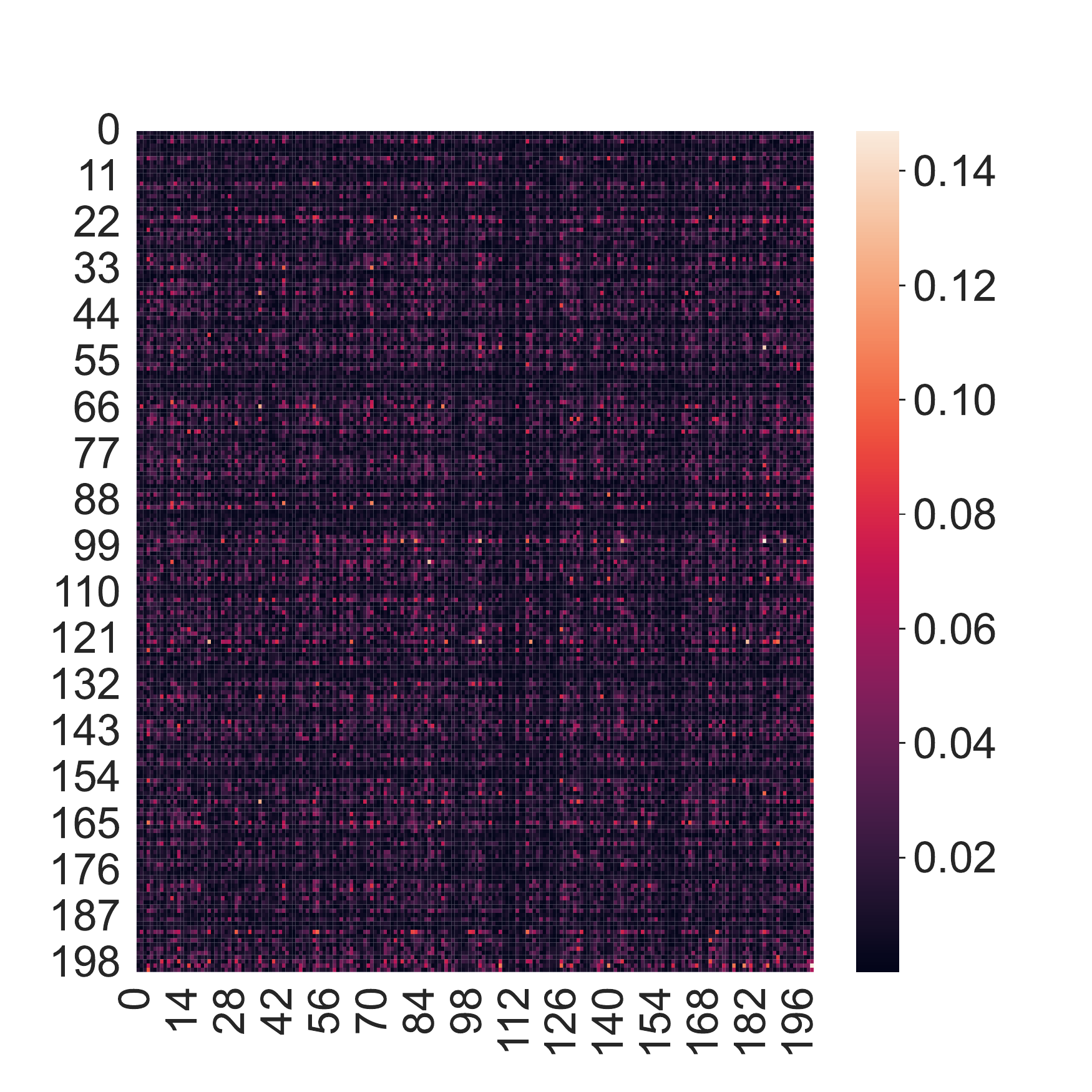}
        \caption{Standard $\mathcal{L}_2$ regularizer}
        \label{fig:exp1-weights-fnn-l2}
    \end{subfigure}
    
    \begin{subfigure}{.5\textwidth}
    \centering
        \includegraphics[scale=0.28]{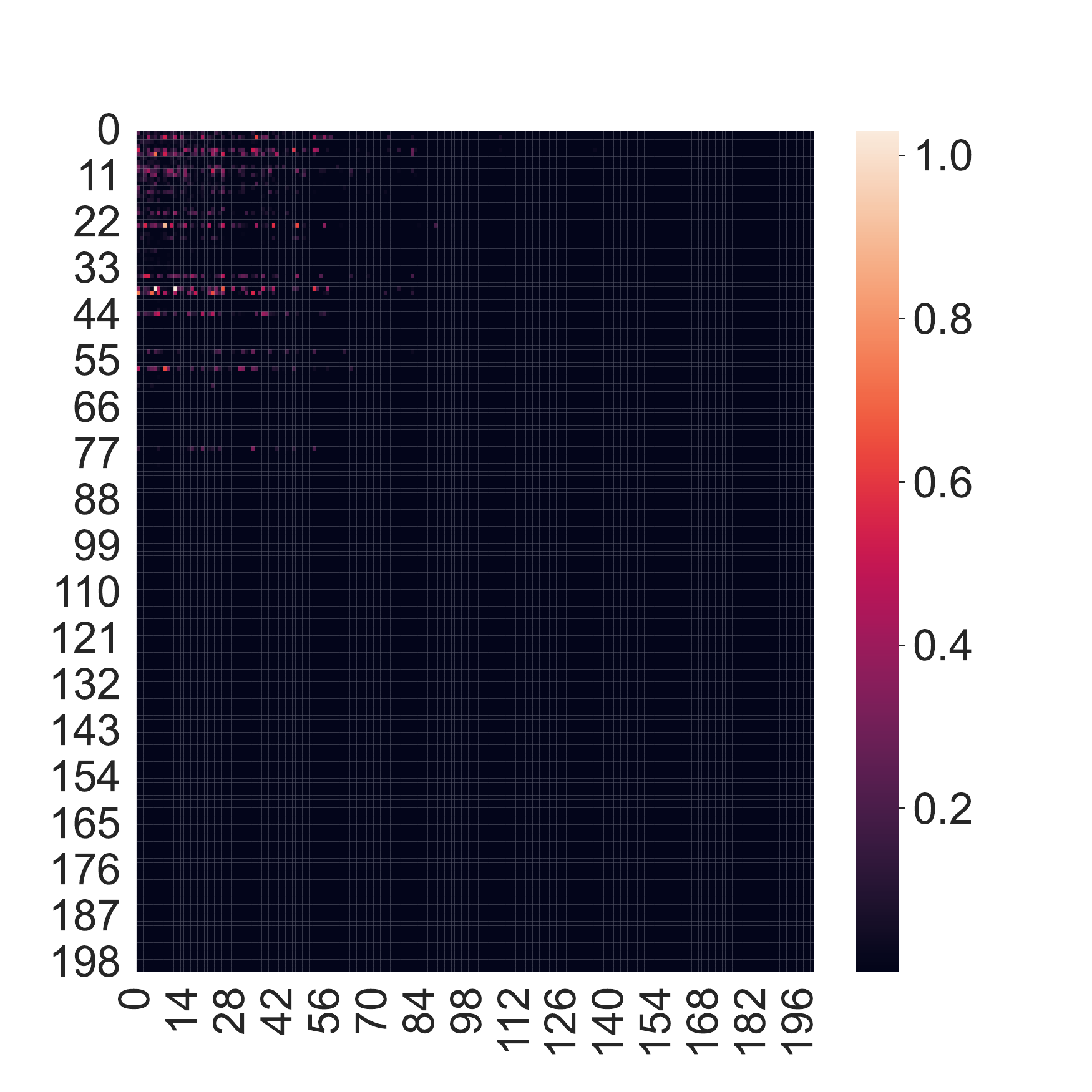}
        \caption{Guided $\mathcal{L}_1$ Regularizer}
        \label{fig:exp1-weights-fnn-gl1}
    \end{subfigure}%
    \begin{subfigure}{.5\textwidth}
    \centering
        \includegraphics[scale=0.28]{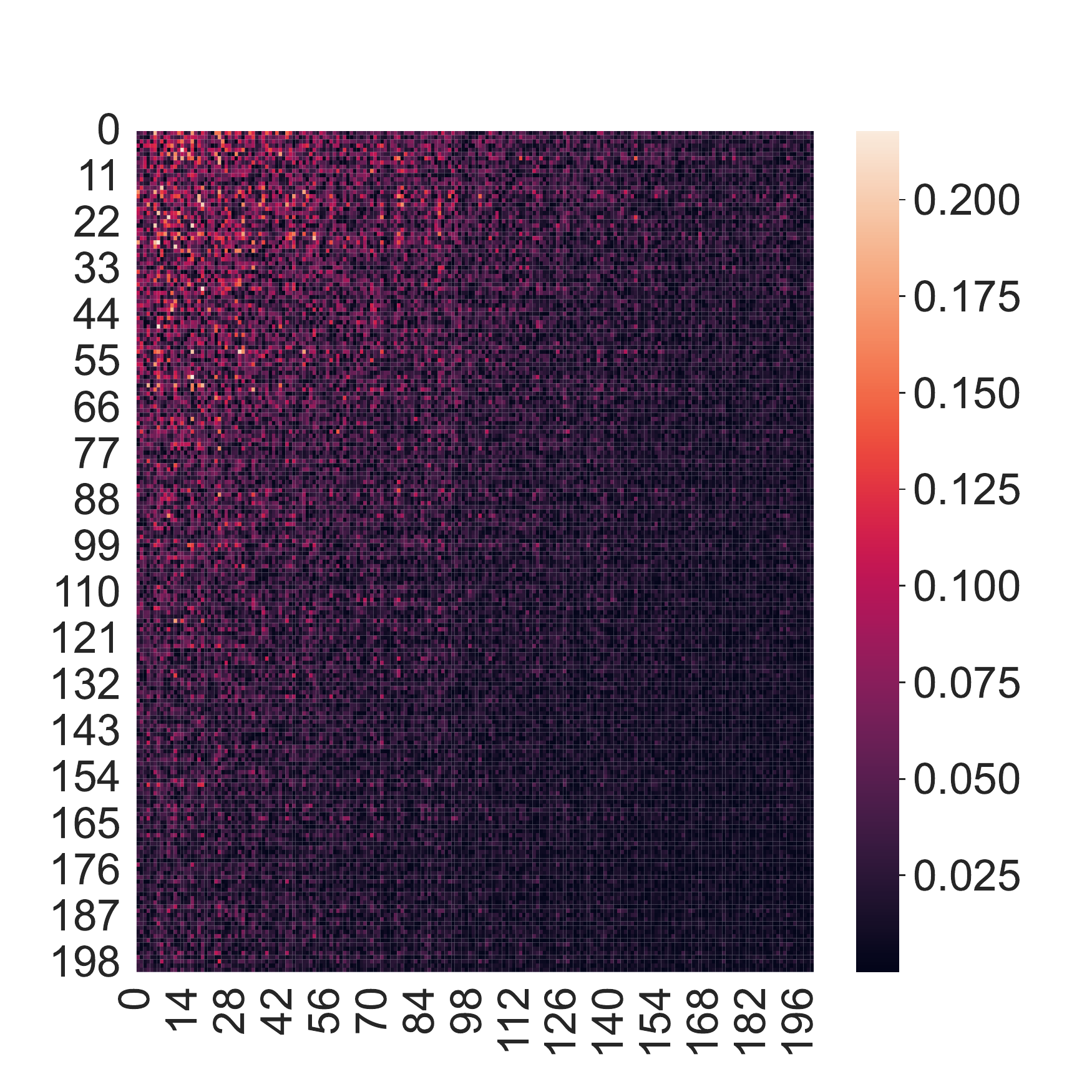}
        \caption{Guided $\mathcal{L}_2$ Regularizer}
        \label{fig:exp1-weights-fnn-gl2}
    \end{subfigure}%
    \caption{Experiment 1. Heatmaps  (absolute value of the elements) for the weight matrices between the two hidden layers of the fully connected network. Five different weight matrix heatmaps are given here for different regularization settings.}
    \label{fig:exp1-weights-fnn}
\end{figure}

\begin{figure}[htb!]
\centering
\begin{subfigure}{0.5\textwidth}
    \centering
    \includegraphics[scale=0.4]{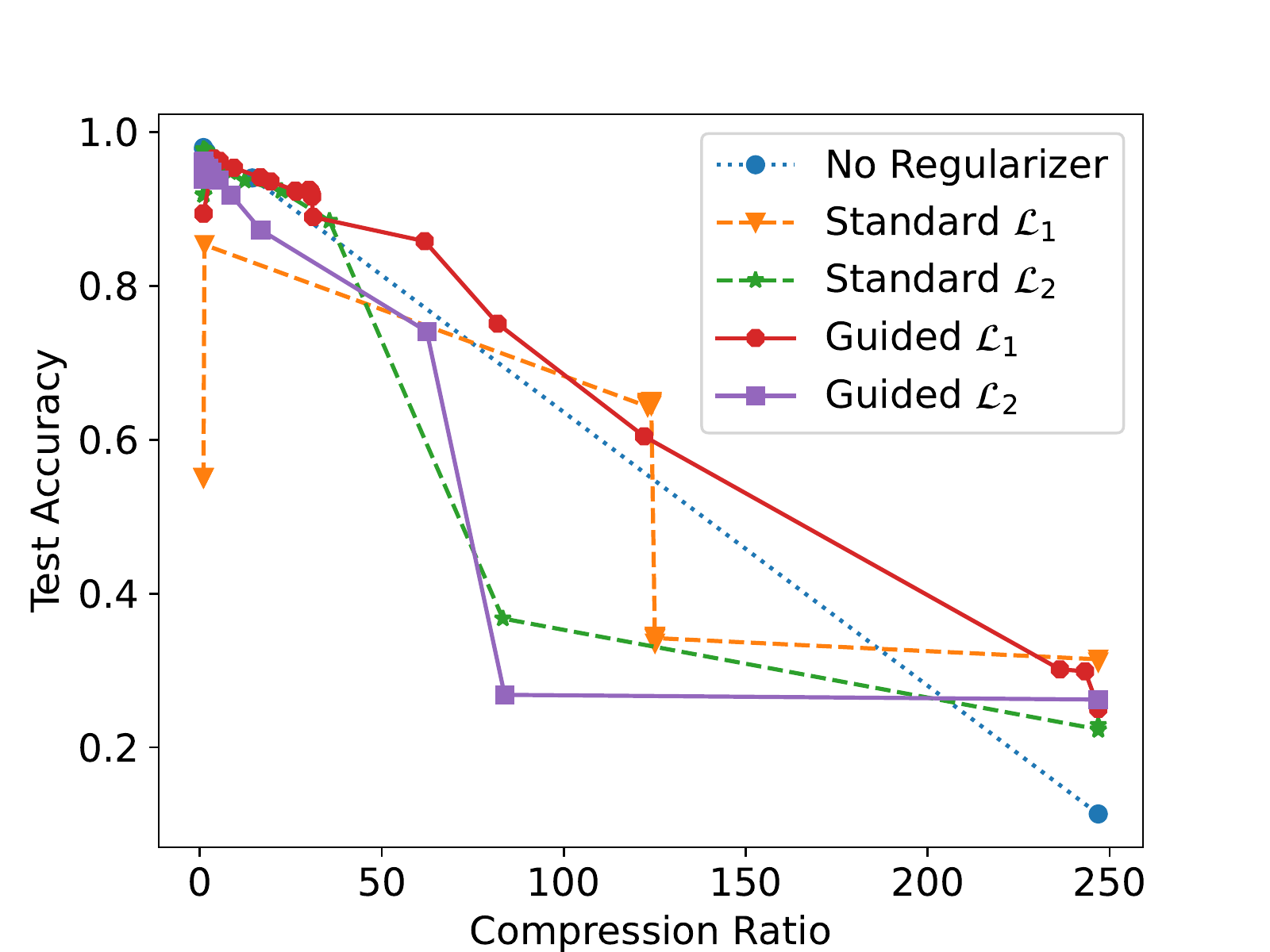}
    \caption{With fine-tuning}
\end{subfigure}%
    \begin{subfigure}{0.5\textwidth}
    \centering
    \includegraphics[scale=0.4]{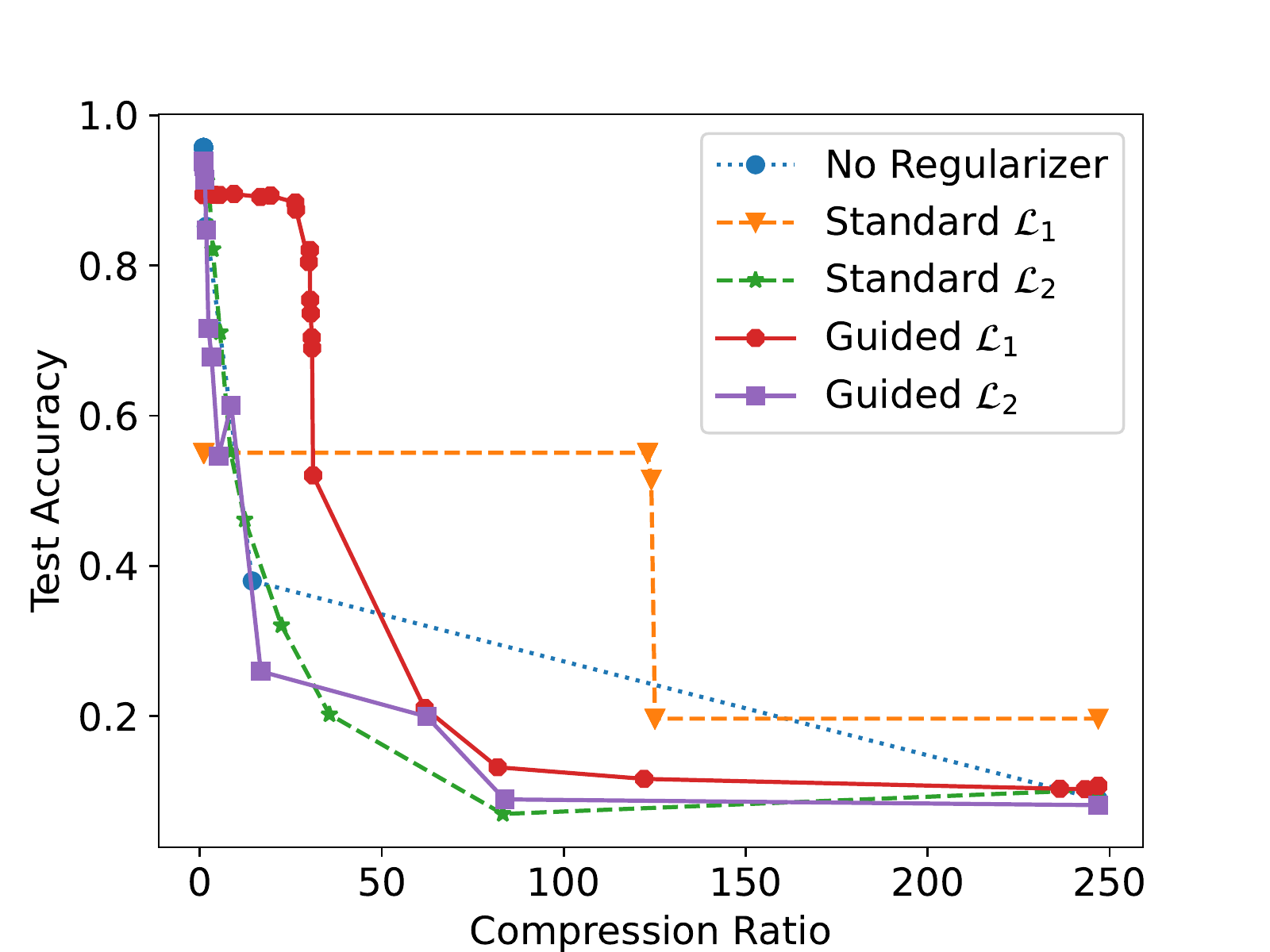}
    \caption{Without fine-tuning}
\end{subfigure}
    \caption{Experiment 1. Test accuracies of the reduced NNs for varying compression ratios with different regularizers.}
    \label{fig:exp1-acc-vs-cr}
\end{figure}

Figure \ref{fig:exp1-weights-fnn} shows the sparsity patterns of the weight matrix between the two hidden layers for different choices of regularizers. Figure \ref{fig:exp1-acc-vs-cr} reports test accuracies of the reduced NNs for various compression ratios with different regularizers.

The weight matrix does not give us any helpful patterns that we can use to prune neurons while using no regularizer (Figure \ref{fig:exp1-weights-fnn-none}), $\mathcal{L}_1$ regularizer (Figure \ref{fig:exp1-weights-fnn-l1}) and $\mathcal{L}_2$ regularizer (Figure \ref{fig:exp1-weights-fnn-l2}). The $\mathcal{L}_1$ regularizer also pushed most of the weight components to a low value (order of $10^{-5}$) for this small network and performs worse than guided regularizers as can be seen in Figure \ref{fig:exp1-acc-vs-cr}. Even though the performance of the model with $\mathcal{L}_1$ regularizer improves considerably after fine-tuning, the predictions are nearly random even for compression ratio = 1. The performance might have improved in this case because we fine-tuned without any regularizers. The guided $\mathcal{L}_2$ regularizer (Figure \ref{fig:exp1-weights-fnn-gl2}) also does not help us as well as guided $\mathcal{L}_1$ in terms of pruning neurons. In the case of guided $\mathcal{L}_1$ regularizer (Figure \ref{fig:exp1-weights-fnn-gl1}), only some of the weight components in the upper-left corner of the matrix have non-zero values. So, visually, we can eliminate a lot of neurons from the hidden layers. From Figure \ref{fig:exp1-acc-vs-cr}, we can see that after pruning, the model trained with guided $\mathcal{L}_1$ performs better than all the other models both before and after fine-tuning.

\subsection{Experiment 2: Comparison Against State-of-the-Art Methods}
\subsubsection{Methods Considered}
We compare our method against five state-of-the-art pruning methods from the literature:
\begin{itemize}
    \item \textbf{LayerInChange \cite{el2022data}:}  This method and the two variants, \textbf{SeqInChange} and \textbf{AsymInChange}, are considered to be the current state-of-the-art for NN pruning. Neurons/channels are pruned in a way that the Frobenius-norm of the difference between ($\text{activations}\times\text{weights}$) before and after pruning is minimized. This is done by formulating the problem as a weak submodular maximization problem and greedily selecting the neurons/channels to keep. In the \textbf{LayerInChange} variant, the layers are pruned independently. In \textbf{SeqInChange}, the layers are pruned sequentially, starting from the first layer. \textbf{AsymInChange} is also sequential pruning but with a defense in place to avoid the accumulation of errors.
    
    \item \textbf{LayerWeightNorm \cite{li2016pruning}:} In each layer, a certain number of neurons/channels are pruned that have the lowest output weights $\mathcal{L}_1$-norm.

    \item \textbf{ActGrad \cite{molchanov2016pruning}:} Neurons/channels with the lowest ($\text{activations}\times\text{gradients}$), averaged over the training data with layerwise $\mathcal{L}_2$-normalization, are pruned. This also has a variant \textbf{LayerActGrad}, where pruning is done separately in each layer based on  ($\text{activations}\times\text{gradients}$).

    \item \textbf{LayerSampling \cite{liebenwein2019provable}:} Layerwise sampling of neurons/channels is done based on the probabilities proportional to the sensitivities based on ($\text{activations}\times\text{weights}$). The neurons/channels that are not sampled are pruned.

    \item \textbf{LayerGreedyFS \cite{ye2020good}:} All the neurons/channels are removed from a layer and then gradually added back based on a greedy selection. The selection depends on the largest decrease in loss achieved. This method has two variants, a limited data variant that uses some of the data and a full data variant (\textbf{LayerGreedyFS-fd}) that uses all the data.
\end{itemize}

\subsubsection{Experimental Setup}
We evaluate the performance of all the methods for three different neural network models. We trained the LeNet5 model \cite{lecun1998gradient} on the MNIST dataset \cite{lecun1998gradient}. We also trained ResNet56 \cite{he2016deep} and the VGG11 \cite{simonyan2014very} on the CIFAR-10 dataset \cite{krizhevsky2009learning}. The computation of gradients and activations used in \textbf{LayerInChange, LayerSampling} and \textbf{ActGrad} is done using four batches of 128 training images, following \cite{el2022data}. We follow the exact experiment description in \cite{el2022data} for a fair comparison. We used our own implementation for LeNet5 and VGG11 architectures. The last two layers of VGG11 were also changed to have 128 neurons each, as done in \cite{el2022data}. For ResNet56, we used the implementation available in ShrinkBench \cite{blalock2020state}.

All the models were trained for 200 epochs with a batch size of 128 and a learning rate of $10^{-3}$. LeNet5 was optimized using Stochastic Gradient Descent (SGD) with Nesterov momentum value of 0.9 \cite{robbins1951stochastic, nesterov1983method}. The optimizer for VGG11 and ResNet56 was Adam \cite{kingma2014adam} with $\beta_1 = 0.9$ and $\beta_2 = 0.99$. After the training and pruning, LeNet5 was fine-tuned for 10 epochs. VGG11 and ResNet56 were fine-tuned for 20 epochs. During fine-tuning, we used the normal loss functions. For all the models, Adam with $\beta_1 = 0.9$ and $\beta_2 = 0.99$ was applied during fine-tuning for optimization. We applied our method for nine values of the regularizer penalty, $\lambda \in \{10^{-8}, 10^{-7}, 10^{-6}, 10^{-5}, 10^{-4}, 10^{-3}, 10^{-2}, 10^{-1}, 1\}$ during training. All the results reported are after pruning and fine-tuning the model. We decided to use only the guided $\mathcal{L}_1$ regularizer for comparison because the guided  $\mathcal{L}_2$ regularizer was not as effective in Experiment 1.

For LeNet5, we pruned all the layers except the output layer. For VGG11, we prune every layer except the output layer and the last convolution layer. For ResNet56, all the layers are pruned except the last layer in each residual block, the last layer before each residual block and the output layer. 

\subsubsection{Results and Discussion}
For comparison with the state-of-the-art methods, we chose the $\lambda$ value to be $10^{-3}$, $10^{-4}$ and $10^{-4}$ for LeNet5, ResNet56, and VGG11, respectively, as our pruning method performed the best for these $\lambda$ values. 

\begin{figure}[htb!]
    \centering
    \begin{subfigure}{.5\textwidth}
    \centering
    \includegraphics[scale=0.4]{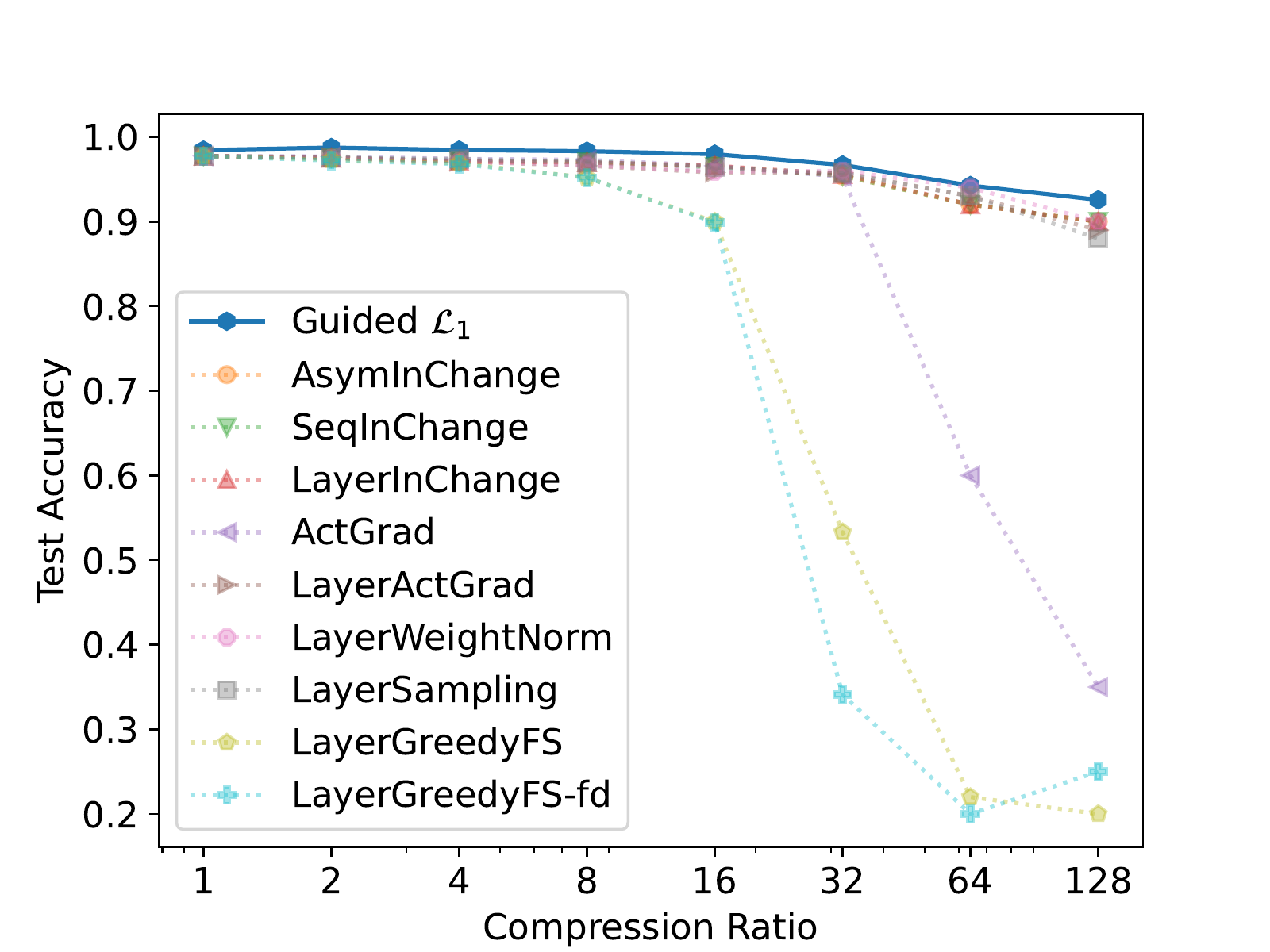}
    \caption{LeNet5}
    \end{subfigure}%
    \begin{subfigure}{.5\textwidth}
    \centering
        \includegraphics[scale=0.4]{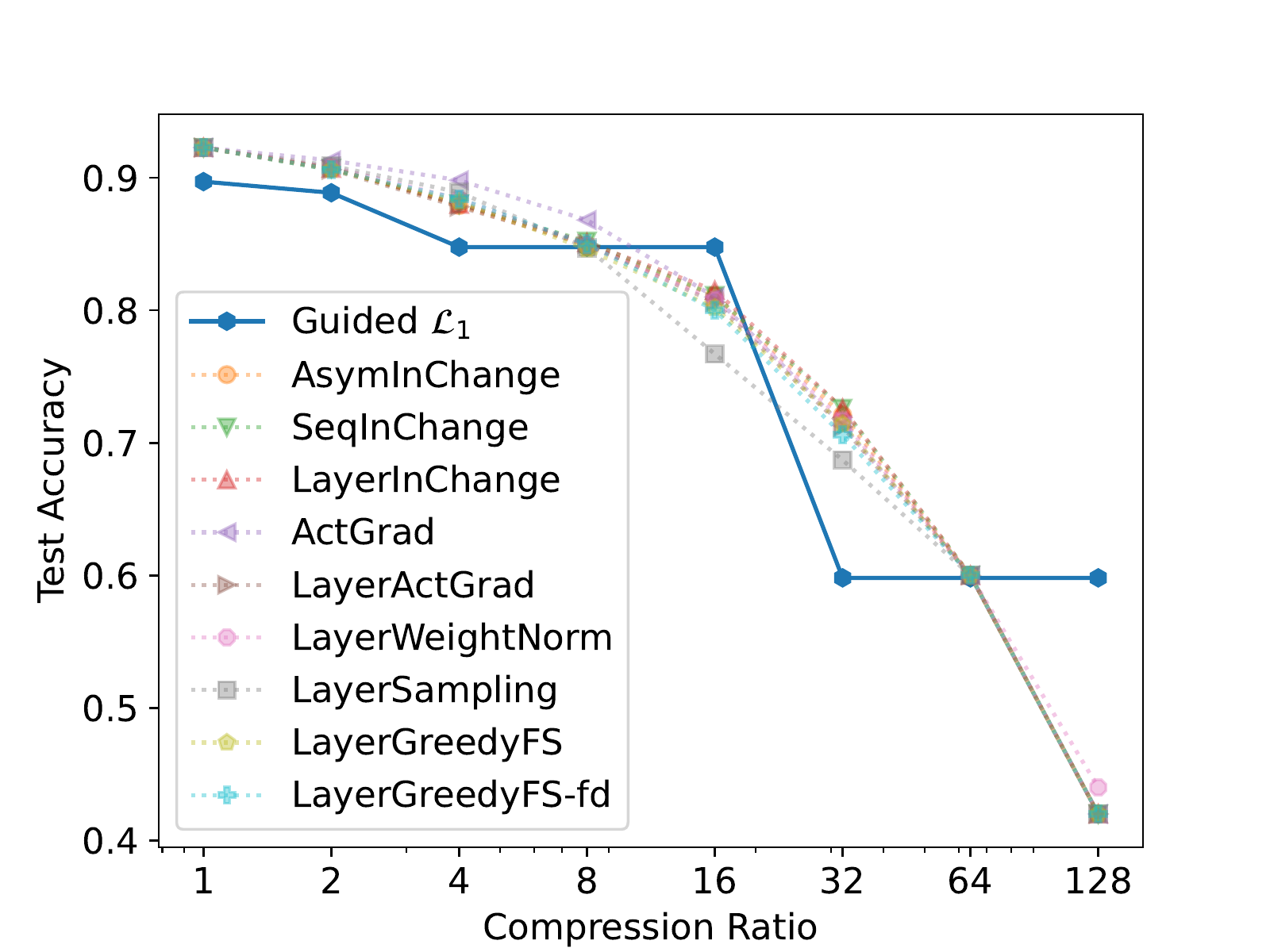}
    \caption{ResNet56}
    \end{subfigure}
    
    \begin{subfigure}{.5\textwidth}
    \centering
    \includegraphics[scale=0.4]{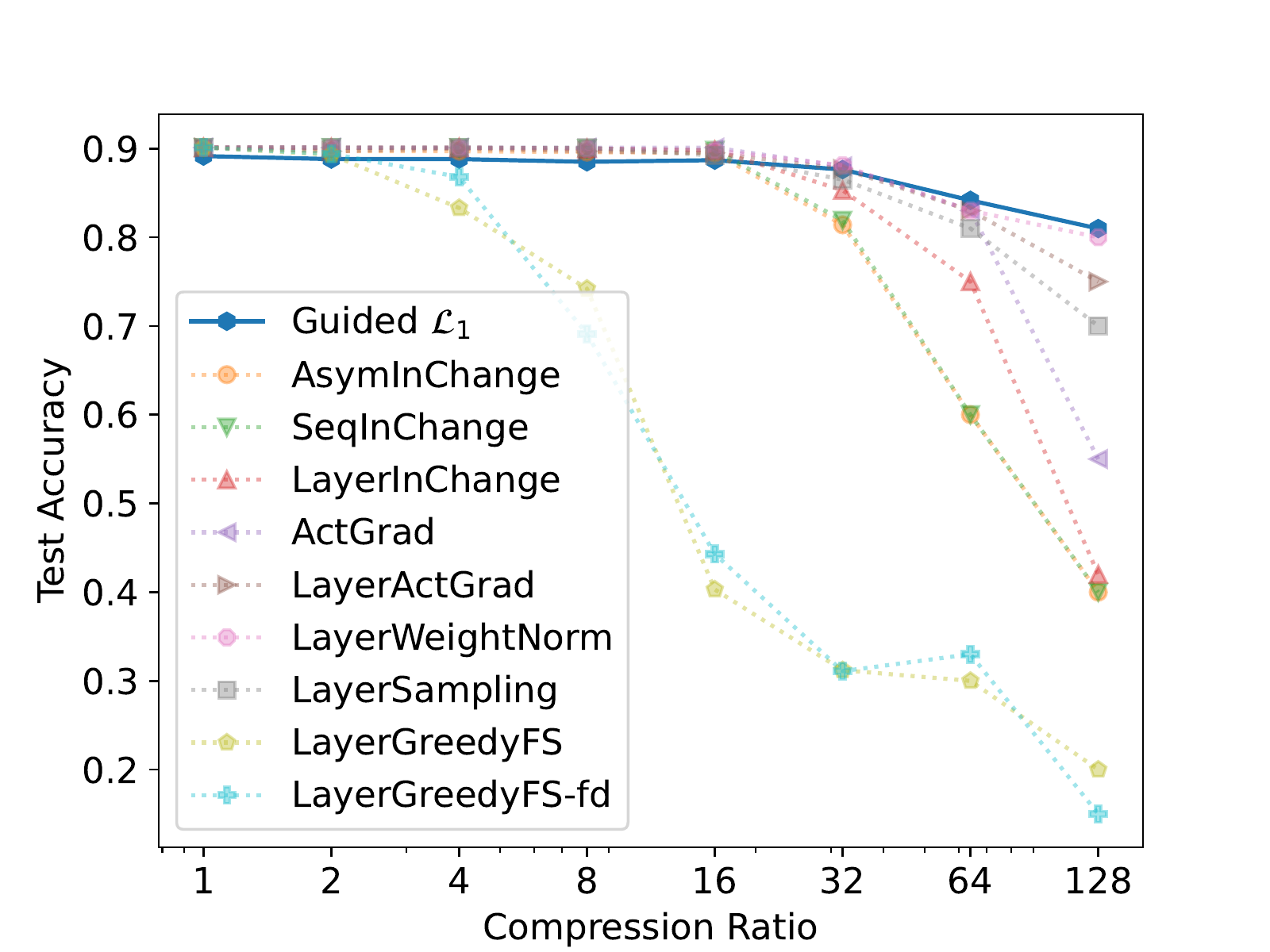}
    \caption{VGG11}
    \end{subfigure}
    \caption{Test accuracies for different pruning methods and varying compression ratios. The guided regularization method proposed herein performs at least as well as any of the existing approaches.}
    \label{fig:exp2-acc-vs-cr-comparisons}
\end{figure}

The results of the comparisons are reported in Figure \ref{fig:exp2-acc-vs-cr-comparisons}. The experiments reveal the following:
\begin{itemize}
    \item Our method outperforms all other pruning methods used for LeNet5 and VGG11 models. For the ResNet56 model, our method gives reasonable results, but it is slightly less effective than the state-of-the-art. We think this is due to the depth of the ResNet56 model. During training with the guided regularizer, the effect of regularizing a layer accumulates for the subsequent layers. Studying this is the topic of future work.

    \item For VGG11 and ResNet56, our model achieved an accuracy of 89.15\% and 89.7\% respectively for compression ratio = 1, which is lower than normal training. After normal training, the test accuracy achieved by VGG11 and ResNet56 is 90.11\% and 92.27\% respectively. This is because the pruning is in effect during training while using the guided regularizer.

    \item Using just one single $\alpha$ value \eqref{eq:threshold} for all the layers was effective for our method. In the literature, it is advised to use per-layer budgeting while selecting how much to prune in each layer \cite{el2022data}. However, for our case, the final threshold value $\tau$ is dependent on both $\alpha$ and 
    $\eta_{\text{max}}$. So, how much we prune in each layer depends on the values of the weight components in each layer.
\end{itemize}

\subsection{Implementation and Resources}
We used TensorFlow \cite{tensorflow2015-whitepaper} for implementing the LeNet5 and VGG11 model. The implementation of ResNet56 is done using PyTorch \cite{paszke2017automatic} as we modified the implementation of ShrinkBench \cite{blalock2020state} which uses PyTorch.  We conducted our experiments on three different machines: i) AMD Ryzen 9 5900HS @ 3.30 GHz with 16 GB RAM, NVIDIA GeForce RTX 3070 with 8 GB memory (used for experiment with LeNet5), ii) Intel Xeon W-2155 @ 3.30GHz with 12 GB RAM, NVIDIA Quadro P4000 with 8 GB memory (used for experiment with VGG11) and iii) Intel Xeon E5-2680v4 @ 2.4GHz with 386 GB RAM, NVIDIA Tesla V100 with 16 GB memory (used for experiment with ResNet56).

%% file: conclusion.tex

This work proposes a modification to the well-known regularization techniques of NNs, that guides the NN to achieve prunable patterns in weight units to effectively reduce the NN dimension while also maintaining reasonable performance. We present experimental results that verify the effectiveness of the new methodology. The pruning is virtually in play during training when using the guided regularizer, and so the actual pruning step after pre-training is simpler and more efficient than the previous methods. The threshold based pruning proposed herein is also weight value dependent and does not require per layer budgeting advised in \cite{el2022data}. However, our method was effective but fell slightly behind other state-of-the-art methods for reducing the deep network ResNet56. Future work will seek to improve our proposed method for pruning truly deep neural networks; one strategy is by automatic weight penalty selection.

%% file: main.bbl
\begin{thebibliography}{36}
\providecommand{\natexlab}[1]{#1}
\providecommand{\url}[1]{\texttt{#1}}
\expandafter\ifx\csname urlstyle\endcsname\relax
  \providecommand{\doi}[1]{doi: #1}\else
  \providecommand{\doi}{doi: \begingroup \urlstyle{rm}\Url}\fi

\bibitem[Abadi et~al.(2015)Abadi, Agarwal, Barham, Brevdo, Chen, Citro,
  Corrado, Davis, Dean, Devin, Ghemawat, Goodfellow, Harp, Irving, Isard, Jia,
  Jozefowicz, Kaiser, Kudlur, Levenberg, Man\'{e}, Monga, Moore, Murray, Olah,
  Schuster, Shlens, Steiner, Sutskever, Talwar, Tucker, Vanhoucke, Vasudevan,
  Vi\'{e}gas, Vinyals, Warden, Wattenberg, Wicke, Yu, and
  Zheng]{tensorflow2015-whitepaper}
Mart\'{i}n Abadi, Ashish Agarwal, Paul Barham, Eugene Brevdo, Zhifeng Chen,
  Craig Citro, Greg~S. Corrado, Andy Davis, Jeffrey Dean, Matthieu Devin,
  Sanjay Ghemawat, Ian Goodfellow, Andrew Harp, Geoffrey Irving, Michael Isard,
  Yangqing Jia, Rafal Jozefowicz, Lukasz Kaiser, Manjunath Kudlur, Josh
  Levenberg, Dandelion Man\'{e}, Rajat Monga, Sherry Moore, Derek Murray, Chris
  Olah, Mike Schuster, Jonathon Shlens, Benoit Steiner, Ilya Sutskever, Kunal
  Talwar, Paul Tucker, Vincent Vanhoucke, Vijay Vasudevan, Fernanda Vi\'{e}gas,
  Oriol Vinyals, Pete Warden, Martin Wattenberg, Martin Wicke, Yuan Yu, and
  Xiaoqiang Zheng.
\newblock {TensorFlow}: Large-scale machine learning on heterogeneous systems,
  2015.
\newblock URL \url{https://www.tensorflow.org/}.
\newblock Software available from tensorflow.org.

\bibitem[Blalock et~al.(2020)Blalock, Gonzalez~Ortiz, Frankle, and
  Guttag]{blalock2020state}
Davis Blalock, Jose~Javier Gonzalez~Ortiz, Jonathan Frankle, and John Guttag.
\newblock What is the state of neural network pruning?
\newblock \emph{Proceedings of machine learning and systems}, 2:\penalty0
  129--146, 2020.

\bibitem[Buciluǎ et~al.(2006)Buciluǎ, Caruana, and
  Niculescu-Mizil]{buciluǎ2006model}
Cristian Buciluǎ, Rich Caruana, and Alexandru Niculescu-Mizil.
\newblock Model compression.
\newblock In \emph{Proceedings of the 12th ACM SIGKDD international conference
  on Knowledge discovery and data mining}, pages 535--541, 2006.

\bibitem[Collins and Kohli(2014)]{collins2014memory}
Maxwell~D Collins and Pushmeet Kohli.
\newblock Memory bounded deep convolutional networks.
\newblock \emph{arXiv preprint arXiv:1412.1442}, 2014.

\bibitem[Courbariaux et~al.(2015)Courbariaux, Bengio, and
  David]{courbariaux2015binaryconnect}
Matthieu Courbariaux, Yoshua Bengio, and Jean-Pierre David.
\newblock Binaryconnect: Training deep neural networks with binary weights
  during propagations.
\newblock \emph{Advances in neural information processing systems}, 28, 2015.

\bibitem[Denil et~al.(2013)Denil, Shakibi, Dinh, Ranzato, and
  De~Freitas]{denil2013predicting}
Misha Denil, Babak Shakibi, Laurent Dinh, Marc'Aurelio Ranzato, and Nando
  De~Freitas.
\newblock Predicting parameters in deep learning.
\newblock \emph{Advances in neural information processing systems}, 26, 2013.

\bibitem[El~Halabi et~al.(2022)El~Halabi, Srinivas, and
  Lacoste-Julien]{el2022data}
Marwa El~Halabi, Suraj Srinivas, and Simon Lacoste-Julien.
\newblock Data-efficient structured pruning via submodular optimization.
\newblock \emph{Advances in Neural Information Processing Systems},
  35:\penalty0 36613--36626, 2022.

\bibitem[Gong et~al.(2014)Gong, Liu, Yang, and Bourdev]{gong2014compressing}
Yunchao Gong, Liu Liu, Ming Yang, and Lubomir Bourdev.
\newblock Compressing deep convolutional networks using vector quantization.
\newblock \emph{arXiv preprint arXiv:1412.6115}, 2014.

\bibitem[Han et~al.(2015)Han, Pool, Tran, and Dally]{han2015learning}
Song Han, Jeff Pool, John Tran, and William Dally.
\newblock Learning both weights and connections for efficient neural network.
\newblock \emph{Advances in neural information processing systems}, 28, 2015.

\bibitem[He et~al.(2016)He, Zhang, Ren, and Sun]{he2016deep}
Kaiming He, Xiangyu Zhang, Shaoqing Ren, and Jian Sun.
\newblock Deep residual learning for image recognition.
\newblock In \emph{Proceedings of the IEEE conference on computer vision and
  pattern recognition}, pages 770--778, 2016.

\bibitem[He et~al.(2014)He, Fan, Qian, Tan, and Yu]{he2014reshaping}
Tianxing He, Yuchen Fan, Yanmin Qian, Tian Tan, and Kai Yu.
\newblock Reshaping deep neural network for fast decoding by node-pruning.
\newblock In \emph{2014 IEEE International Conference on Acoustics, Speech and
  Signal Processing (ICASSP)}, pages 245--249. IEEE, 2014.

\bibitem[He et~al.(2017)He, Zhang, and Sun]{he2017channel}
Yihui He, Xiangyu Zhang, and Jian Sun.
\newblock Channel pruning for accelerating very deep neural networks.
\newblock In \emph{Proceedings of the IEEE international conference on computer
  vision}, pages 1389--1397, 2017.

\bibitem[Hinton et~al.(2015)Hinton, Vinyals, and Dean]{hinton2015distilling}
Geoffrey Hinton, Oriol Vinyals, and Jeff Dean.
\newblock Distilling the knowledge in a neural network.
\newblock \emph{arXiv preprint arXiv:1503.02531}, 2015.

\bibitem[Hoefler et~al.(2021)Hoefler, Alistarh, Ben-Nun, Dryden, and
  Peste]{hoefler2021sparsity}
Torsten Hoefler, Dan Alistarh, Tal Ben-Nun, Nikoli Dryden, and Alexandra Peste.
\newblock Sparsity in deep learning: Pruning and growth for efficient inference
  and training in neural networks.
\newblock \emph{The Journal of Machine Learning Research}, 22\penalty0
  (1):\penalty0 10882--11005, 2021.

\bibitem[Hoerl and Kennard(1970)]{hoerl1970ridge}
Arthur~E Hoerl and Robert~W Kennard.
\newblock Ridge regression: Biased estimation for nonorthogonal problems.
\newblock \emph{Technometrics}, 12\penalty0 (1):\penalty0 55--67, 1970.

\bibitem[Kingma and Ba(2014)]{kingma2014adam}
Diederik~P Kingma and Jimmy Ba.
\newblock Adam: A method for stochastic optimization.
\newblock \emph{arXiv preprint arXiv:1412.6980}, 2014.

\bibitem[Krizhevsky et~al.(2009)Krizhevsky, Hinton,
  et~al.]{krizhevsky2009learning}
Alex Krizhevsky, Geoffrey Hinton, et~al.
\newblock Learning multiple layers of features from tiny images.
\newblock 2009.

\bibitem[Kuzmin et~al.(2019)Kuzmin, Nagel, Pitre, Pendyam, Blankevoort, and
  Welling]{kuzmin2019taxonomy}
Andrey Kuzmin, Markus Nagel, Saurabh Pitre, Sandeep Pendyam, Tijmen
  Blankevoort, and Max Welling.
\newblock Taxonomy and evaluation of structured compression of convolutional
  neural networks.
\newblock \emph{arXiv preprint arXiv:1912.09802}, 2019.

\bibitem[Lebedev et~al.(2014)Lebedev, Ganin, Rakhuba, Oseledets, and
  Lempitsky]{lebedev2014speeding}
Vadim Lebedev, Yaroslav Ganin, Maksim Rakhuba, Ivan Oseledets, and Victor
  Lempitsky.
\newblock Speeding-up convolutional neural networks using fine-tuned
  cp-decomposition.
\newblock \emph{arXiv preprint arXiv:1412.6553}, 2014.

\bibitem[LeCun et~al.(1998)LeCun, Bottou, Bengio, and
  Haffner]{lecun1998gradient}
Yann LeCun, L{\'e}on Bottou, Yoshua Bengio, and Patrick Haffner.
\newblock Gradient-based learning applied to document recognition.
\newblock \emph{Proceedings of the IEEE}, 86\penalty0 (11):\penalty0
  2278--2324, 1998.

\bibitem[Li et~al.(2016)Li, Kadav, Durdanovic, Samet, and Graf]{li2016pruning}
Hao Li, Asim Kadav, Igor Durdanovic, Hanan Samet, and Hans~Peter Graf.
\newblock Pruning filters for efficient convnets.
\newblock \emph{arXiv preprint arXiv:1608.08710}, 2016.

\bibitem[Liebenwein et~al.(2019)Liebenwein, Baykal, Lang, Feldman, and
  Rus]{liebenwein2019provable}
Lucas Liebenwein, Cenk Baykal, Harry Lang, Dan Feldman, and Daniela Rus.
\newblock Provable filter pruning for efficient neural networks.
\newblock \emph{arXiv preprint arXiv:1911.07412}, 2019.

\bibitem[Luo et~al.(2017)Luo, Wu, and Lin]{luo2017thinet}
Jian-Hao Luo, Jianxin Wu, and Weiyao Lin.
\newblock Thinet: A filter level pruning method for deep neural network
  compression.
\newblock In \emph{Proceedings of the IEEE international conference on computer
  vision}, pages 5058--5066, 2017.

\bibitem[Mariet and Sra(2015)]{mariet2015diversity}
Zelda Mariet and Suvrit Sra.
\newblock Diversity networks: Neural network compression using determinantal
  point processes.
\newblock \emph{arXiv preprint arXiv:1511.05077}, 2015.

\bibitem[Molchanov et~al.(2016)Molchanov, Tyree, Karras, Aila, and
  Kautz]{molchanov2016pruning}
Pavlo Molchanov, Stephen Tyree, Tero Karras, Timo Aila, and Jan Kautz.
\newblock Pruning convolutional neural networks for resource efficient
  inference.
\newblock \emph{arXiv preprint arXiv:1611.06440}, 2016.

\bibitem[Mussay et~al.(2019)Mussay, Osadchy, Braverman, Zhou, and
  Feldman]{mussay2019data}
Ben Mussay, Margarita Osadchy, Vladimir Braverman, Samson Zhou, and Dan
  Feldman.
\newblock Data-independent neural pruning via coresets.
\newblock \emph{arXiv preprint arXiv:1907.04018}, 2019.

\bibitem[Mussay et~al.(2021)Mussay, Feldman, Zhou, Braverman, and
  Osadchy]{mussay2021data}
Ben Mussay, Dan Feldman, Samson Zhou, Vladimir Braverman, and Margarita
  Osadchy.
\newblock Data-independent structured pruning of neural networks via coresets.
\newblock \emph{IEEE Transactions on Neural Networks and Learning Systems},
  33\penalty0 (12):\penalty0 7829--7841, 2021.

\bibitem[Nesterov(1983)]{nesterov1983method}
Yurii~Evgen'evich Nesterov.
\newblock A method of solving a convex programming problem with convergence
  rate o$\backslash$bigl(k\^{}2$\backslash$bigr).
\newblock In \emph{Doklady Akademii Nauk}, volume 269, pages 543--547. Russian
  Academy of Sciences, 1983.

\bibitem[Paszke et~al.(2017)Paszke, Gross, Chintala, Chanan, Yang, DeVito, Lin,
  Desmaison, Antiga, and Lerer]{paszke2017automatic}
Adam Paszke, Sam Gross, Soumith Chintala, Gregory Chanan, Edward Yang, Zachary
  DeVito, Zeming Lin, Alban Desmaison, Luca Antiga, and Adam Lerer.
\newblock Automatic differentiation in pytorch.
\newblock 2017.

\bibitem[Robbins and Monro(1951)]{robbins1951stochastic}
Herbert Robbins and Sutton Monro.
\newblock A stochastic approximation method.
\newblock \emph{The annals of mathematical statistics}, pages 400--407, 1951.

\bibitem[Simonyan and Zisserman(2014)]{simonyan2014very}
Karen Simonyan and Andrew Zisserman.
\newblock Very deep convolutional networks for large-scale image recognition.
\newblock \emph{arXiv preprint arXiv:1409.1556}, 2014.

\bibitem[Srinivas and Babu(2015)]{srinivas2015data}
Suraj Srinivas and R~Venkatesh Babu.
\newblock Data-free parameter pruning for deep neural networks.
\newblock \emph{arXiv preprint arXiv:1507.06149}, 2015.

\bibitem[Su et~al.(2018)Su, Li, Bhattacharjee, and Huang]{su2018tensorial}
Jiahao Su, Jingling Li, Bobby Bhattacharjee, and Furong Huang.
\newblock Tensorial neural networks: Generalization of neural networks and
  application to model compression.
\newblock \emph{arXiv preprint arXiv:1805.10352}, 2018.

\bibitem[Tibshirani(1996)]{tibshirani1996regression}
Robert Tibshirani.
\newblock Regression shrinkage and selection via the lasso.
\newblock \emph{Journal of the Royal Statistical Society: Series B
  (Methodological)}, 58\penalty0 (1):\penalty0 267--288, 1996.

\bibitem[Voita et~al.(2019)Voita, Talbot, Moiseev, Sennrich, and
  Titov]{voita2019analyzing}
Elena Voita, David Talbot, Fedor Moiseev, Rico Sennrich, and Ivan Titov.
\newblock Analyzing multi-head self-attention: Specialized heads do the heavy
  lifting, the rest can be pruned.
\newblock \emph{arXiv preprint arXiv:1905.09418}, 2019.

\bibitem[Ye et~al.(2020)Ye, Gong, Nie, Zhou, Klivans, and Liu]{ye2020good}
Mao Ye, Chengyue Gong, Lizhen Nie, Denny Zhou, Adam Klivans, and Qiang Liu.
\newblock Good subnetworks provably exist: Pruning via greedy forward
  selection.
\newblock In \emph{International Conference on Machine Learning}, pages
  10820--10830. PMLR, 2020.

\end{thebibliography}
